\documentclass[conference]{IEEEtran}
\IEEEoverridecommandlockouts

\usepackage{cite}
\usepackage{amsmath,amssymb,amsfonts}
\usepackage{algorithmic}
\usepackage{graphicx}
\usepackage{textcomp}
\usepackage[dvipsnames]{xcolor}
\usepackage{subcaption}
\usepackage{url}
\usepackage{hyperref}
\usepackage{adjustbox}
\usepackage{todonotes}

\begin{document}

\title{Deep Speech Representations Track Word Timing along a Low-Dimensional Circular Trajectory
}

\author{\IEEEauthorblockN{Burin Naowarat, Hao Tang, Sharon Goldwater}
\IEEEauthorblockA{\textit{The Centre for Speech Technology Research,
School of Informatics,
University of Edinburgh}\\
burin.naowarat@ed.ac.uk, hao.tang@ed.ac.uk, sgwater@inf.ed.ac.uk}
}

\maketitle

\begin{abstract}
Speech timing is fundamental for speech models to recognize words accurately, but how word-timing information is represented in speech models remains underexplored. 
We study this question using the framework of representational geometry, 
which investigates whether and how concepts are represented in low-dimensional subspaces of the representation space.
Prior work in cognitive neuroscience has shown that
when HuBERT feature sequences are averaged across words of the same duration, they form near-circular trajectories from word onset to word offset in a two-dimensional ``temporal subspace.'' 
Here we show that this circular structure of word timing is a general property of deep speech representations, by demonstrating its emergence in nine additional speech models covering both supervised and self-supervised. 
We then move beyond averaged word segments and show that in HuBERT, the circular trajectories are consistent enough at the level of individual words to perform word segmentation. Our approach, Rotary Segmentation (RoSeg), achieves 64.3\% boundary F-score on Librispeech by detecting word boundaries when cycles complete in the temporal subspace. Although supervision is needed to identify a good subspace, nevertheless this result demonstrates much stronger implicit word knowledge in HuBERT than has previously been shown. 
\end{abstract}


\IEEEpeerreviewmaketitle

\section{Introduction}

Modeling how speech unfolds over time is a fundamental challenge for speech technology. The success of deep speech models suggests that they must be representing many aspects of speech timing, yet the question of {\em how} they do so has been largely neglected by speech and language technology researchers, despite work investigating many other aspects of speech representations \cite{Pasad2023-zm,pasad_comparative_2023,belinkov_analyzing_2017,martin_probing_2023,shen_encoding_2024,liu_self-supervised_2023,rensburg_interpreting_2026,kloots_tracking_2026,choi_self-supervised_2024,de_seyssel_probing_2022}.

\begin{figure}[t]
    \centering
\begin{subfigure}[t]{\linewidth}
    \adjustbox{valign=t}{\begin{minipage}[t]{0.04\linewidth}
    \hspace{0mm}(a)
    \end{minipage}}
    \adjustbox{valign=t}{\begin{minipage}[t]{0.95\linewidth}
    \includegraphics[width=8.2cm]{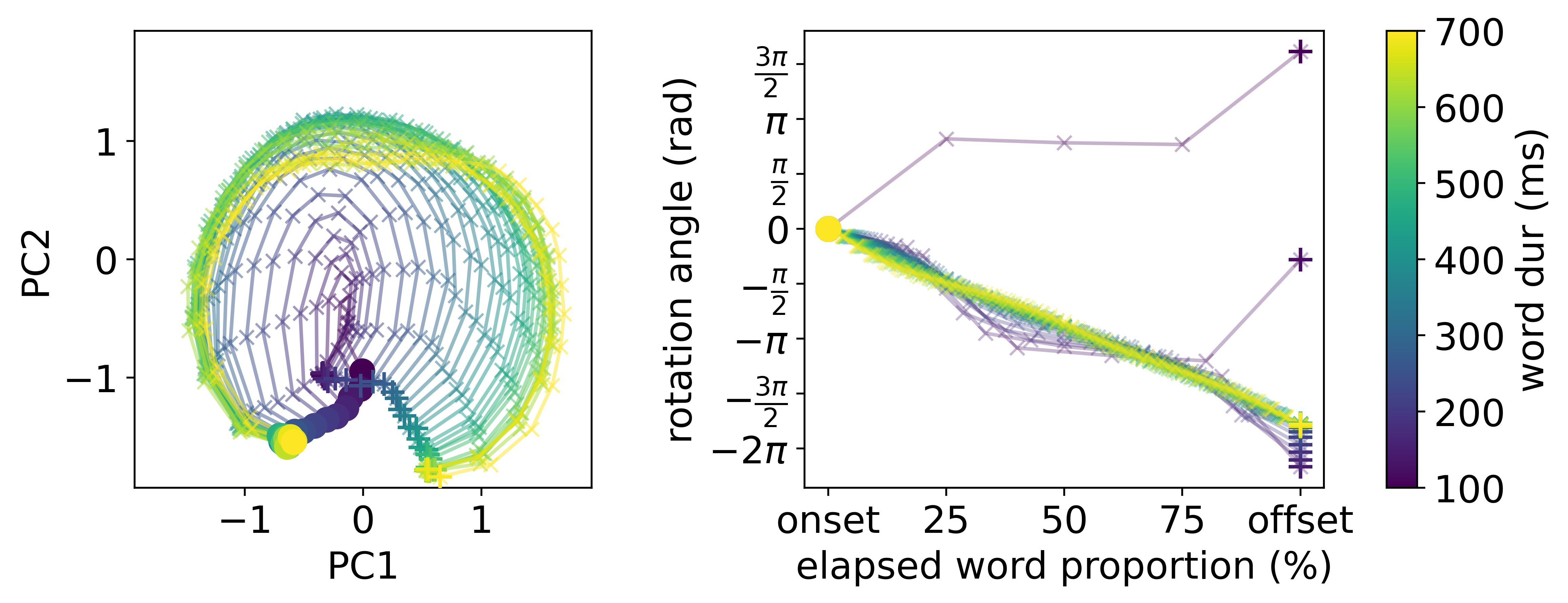}
    \end{minipage}}
    \phantomcaption
    \label{fig:dynamics:word_2d}
\end{subfigure}
\begin{subfigure}[t]{\linewidth}
    \adjustbox{valign=t}{\begin{minipage}[t]{0.04\linewidth}
    \hspace{0mm}(b)
    \end{minipage}}
    \adjustbox{valign=t}{\begin{minipage}[t]{0.95\linewidth}
    \includegraphics[width=8cm]{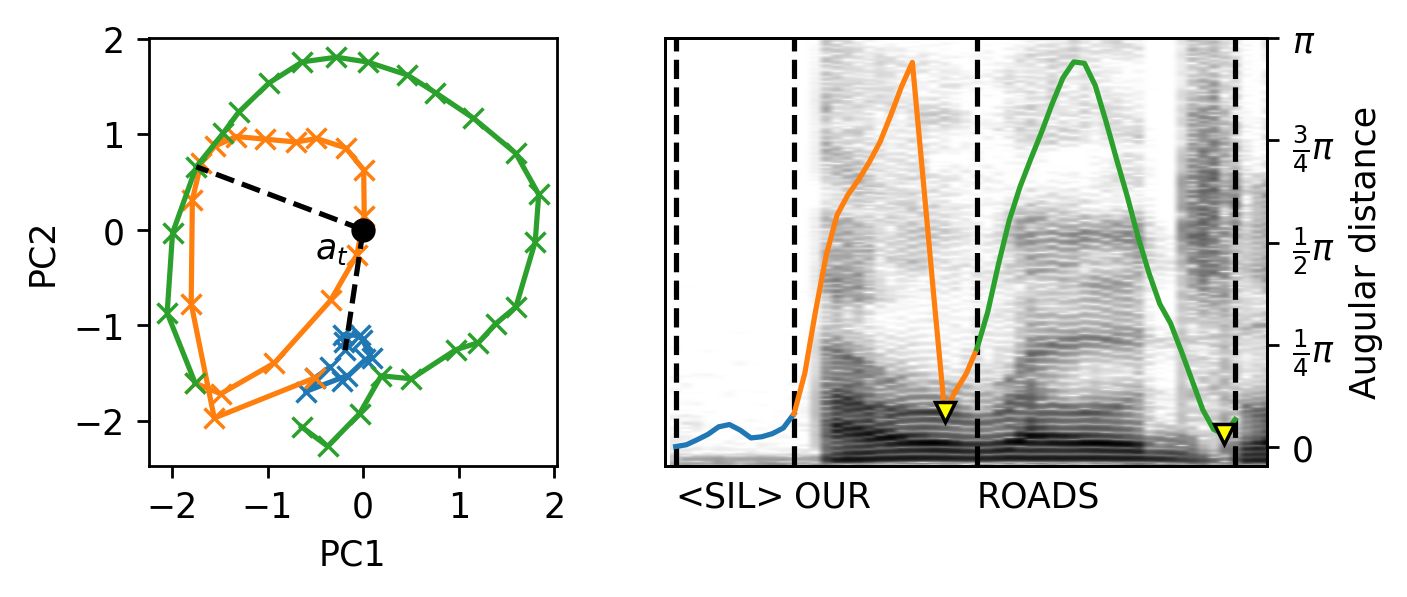}
    \end{minipage}}
    \phantomcaption
    \label{fig:dynamics:instance}
\end{subfigure}
\vspace{-2mm}
    \caption{
    Circular word trajectories track word timing in HuBERT representations.  (a) The two-dimensional projections of averaged word segments onto a temporal subspace reveal circular trajectories. Each trajectory (left) shows the average sequence of HuBERT (layer 9) features in word segments of a particular duration (colours). For for all durations, the average features 
    at each timestep (crosses) follow a nearly complete rotation from word onset (circles) to word offset (pluses).
    The average features for each word duration rotate with roughly constant angular velocity, resulting in a near-linear relationship between the elapsed \% of the word and the rotation angle (right). 
    (b) Frames from individual words (colours) also follow roughly circular trajectories through the temporal subspace (left), so we can
    segment words by predicting boundaries (triangles) at minima of $a_t$, the angular distance relative to utterance onset (right; true boundaries shown as dashed lines). 
    } \vspace{-2mm}
    \label{fig:dynamics}
\end{figure}

In this paper, we specifically examine timing within words. 
We take inspiration from work in cognitive neuroscience, where researchers analyze brain recordings to understand how the brain represents task-relevant information. 
One approach, which we adopt here, is {\em geometric} \cite{yuste_neuron_2015,saxena_towards_2019,gurnani_signatures_2023}: electrode activations recorded over time are treated as a timeseries of high dimensional vector representations (analogous to frame-level representations extracted from a deep speech model), and analyses look for low-dimensional subspaces that correlate with known features of the task or perceptual input. Activations within those subspaces can be traced over time to reveal trajectories associated with variables of interest
\cite{churchland_neural_2012,elsayed_reorganization_2016,Zhang2026-ts, Stephen2023-cv}.
We build on the results of Zhang et al.\ \cite{Zhang2026-ts}, who analyzed brain recordings of humans listening to speech and found
that when the brain representations were averaged across words and projected onto a two-dimensional ``temporal subspace'', the activation patterns followed a near-circular trajectory over time, completing roughly a full rotation between word onset and word offset. They also found a similar pattern in representations extracted from the HuBERT self-supervised speech model \cite{Hsu2021-qo}, as illustrated by our reproduction in Fig.~\ref{fig:dynamics:word_2d}.

Here, we ask two follow-up questions. First, how general are these rotational dynamics? Do they emerge in other deep speech representations, regardless of model architecture or training paradigm? If so (and if they are not present in more naive representations of the input signal), this suggests that they may be a very general solution to the important problem of tracking speech over time. 
We address this question by analysing representations extracted from English LibriSpeech data by ten different deep speech models. We found that similar patterns emerge across models with diverse 
architectures and training objectives, including both self-supervised and supervised models; causal and non-causal architectures; and various positional encodings.
Thus, circular trajectories in averaged word representations appear to be a very general property of deep speech representations.

Second, if these circular trajectories are a meaningful aspect of speech representations, we would expect them to track word timing not just {\em on average}, but also at the level of individual words. We investigate whether this is the case by proposing Rotary Segmentation (RoSeg), a novel word segmentation algorithm based on rotational dynamics. RoSeg projects the speech representations into a two-dimensional temporal subspace and then predicts a word boundary whenever a full rotation completes. In experiments using HuBERT, we explore two different methods for estimating the temporal subspace and find that with the better method, RoSeg achieves F-scores of 64.3\% (boundaries) and 39.8\% (tokens) on LibriSpeech test-clean, with only moderate degradation on multi-syllable words. This suggests that even in a self-supervised model trained without knowledge of words, the learned representations are tracking word instances in a low-dimensional temporal subspace. Our results provide evidence of stronger implicit word knowledge in HuBERT than has previously been demonstrated: for example, previous papers showed that peaks in dissimilarity between adjacent representation frames tend to indicate word boundaries \cite{Pasad2023-zm,Malan2024-eo}, but using the best published method to segment at these peaks only achieved boundary and token F-scores of 43.9\% and 16.9\%, respectively.

Overall, our findings provide novel insights into how speech models represent word timing, and demonstrate that self-supervision alone can provide models with considerable implicit knowledge of wordforms. 
More broadly, our work exemplifies how a geometric approach can complement typical probing-style analyses \cite{Pasad2023-zm,pasad_comparative_2023,belinkov_analyzing_2017,cho_evidence_2023,martin_probing_2023,shen_encoding_2024}, providing information about how information is structured in the representations, not just how accessible it is. While geometric analyses and interventions have been widely adopted in NLP (e.g., \cite{Belrose2023-ny,hernandez_low-dimensional_2021,Chang2022-ky,park_linear_2023,zou_representation_2023,geiger_finding_2024}), this type of approach is still rare in speech (though see \cite{liu_self-supervised_2023,choi_self-supervised_2026,Naowarat2026-lt,stephenson_untangling_2019,abdullah_analyzing_2022}), and our work illustrates how it can answer fundamental questions unique to this domain.

\section{Circular trajectories tracking word timing can be found in a wide range of models
}
\label{sec:emergent_dynamics}

This section first explains our methods, and then demonstrates that the circular patterns identified in pre-trained HuBERT \cite{Zhang2026-ts} also emerge in representations from diverse speech models. 
We do not find these patterns in an untrained ASR model or in log Mel representations, but we do find a weaker form of the same pattern in an untrained HuBERT model. We discuss possible reasons for this last finding, and investigate these further in Section~\ref{sec:segmentation}.

\subsection{Temporal subspace estimation}
\label{sec:temporal_subspaces}

We define a {\em word temporal subspace} (or simply {\em temporal subspace}) as a low-dimensional subspace of the representation space that encodes word timing information. 
Identifying such a subspace is an example of a {\em concept estimation} problem, where the concept is word timing. 

Recent work \cite{Naowarat2026-lt} has highlighted that different estimators may identify distinct subspaces for the same concept, depending on their objective function. 
For simplicity, in this section we use the same estimator as in \cite{Zhang2026-ts}, which we refer to as Centroid-PCA (CPCA), following \cite{Naowarat2026-lt}. Section~\ref{sec:segmentation} explores an alternative estimator which we find works better at the individual word level. Note that both estimators are supervised: they require knowledge of word boundaries in order to estimate the temporal subspace.

CPCA estimates the word temporal subspace as follows. Given a sequence of $N$-dimensional representations extracted from the speech input and the corresponding word boundaries,
the word segments are grouped by duration and averaged across all words within each group. For a group with duration $T$, this yields a sequence $\bar{x}_1 \ldots \bar{x}_T$ of average 
frames, where each $\bar{x}_t \in \mathbb{R}^N$ is the average of
the $t$-th frame over all word segments of duration $T$.
Variation due to non-timing factors is averaged out, so variation between the centroids 
is associated with word timing information. PCA is then applied to the collection of all centroids across all duration groups to estimate a low-dimensional subspace that maximally preserves the variance associated with word timing.\footnote{CPCA as defined by \cite{liu_self-supervised_2023} and studied in \cite{Naowarat2026-lt} simply computes PCA over the centroids of the classes of interest (in their case, phone or speaker labels). 
Arguably here the classes of interest are just the time steps relative to word onset, but grouping the words by duration first can also be viewed as CPCA, where the classes are pairs of (timestep, word duration) labels. In practice, we found that including the duration grouping as in \cite{Zhang2026-ts} worked better, both in this section and for word segmentation (Section~\ref{sec:segmentation}).
} 
To analyze word circles, we consider only the top two components of the PCA, which account for between 54\% and 79\% of the variance in the centroids computed from any of the trained models/layers we tested. Future work could explore the remaining dimensions.

\begin{table*}[]
\caption{
Models used in our first study. 
We abbreviate Trans(formers), Conv(olutional), positional encoding (PE), and LibriSpeech (LS).
All models have a frame rate of 50 Hz except Mimi and Vibevoice, whose frame rates vary across layers. 
For Mimi, we used the 25 Hz continuous features from the Transformer layer. For Vibevoice, we used layer 5, the only one with a similar frame rate to other models  (60 Hz).
}

\hspace{-6mm}
\begin{tabular}{lllllll}
\hline
Model              & Architecture     & Training paradigm       & Prediction target & PE     & Dataset     & HuggingFace name               \\
\hline
HuBERT \cite{Hsu2021-qo}            & Non-causal Trans & Masked prediction & Cluster ID         & ConvPE & LS 960& facebook/hubert-base-ls960 \\
MelHuBERT \cite{Lin2023-aa}         & Non-causal Trans & Masked prediction & Cluster ID         & ConvPE & LS 360      & -                          \\
Wav2Vec2 \cite{Baevski2020-vc}          & Non-causal Trans & Masked prediction & Online codebook           & ConvPE & LS 960      & facebook/wav2vec2-base     \\
WavLM \cite{Chen2021-fe}              & Non-causal Trans & Masked prediction + denoising & Cluster ID         & ConvPE & LS 960      & microsoft/wavlm-base       \\
DinoSR \cite{Liu2023-lh}             & Non-causal Trans & Masked prediction + distillation & Online cluster ID  & ConvPE & LS 960      & -                          \\
SpidR \cite{Poli2025-kv}             & Non-causal Trans & Masked prediction + distillation & Online cluster ID  & ConvPE & LS 960      & -                          \\
Whisper encoder \cite{Radford2023-ef}   & Non-causal Trans & ASR                     & Transcripts        & SinePE & 680k hours  & openai/whisper-base        \\
Mimi \cite{Defossez2024-ce}              & Causal Trans     & Speech codec + distillation   & Speech             & RoPE   & 7M hours    & kyutai/mimi                \\
VibeVoice-acoustic \cite{Peng2025-bi} & Causal Conv      & Variational AutoEncoder & Speech              & No PE   & Unspecified & microsoft/VibeVoice-ASR-HF \\
VibeVoice-semantic \cite{Peng2025-bi} & Causal Conv      & ASR                     & Transcripts        & No PE   & Unspecified & microsoft/VibeVoice-ASR-HF \\
\hline

\end{tabular}
\vspace{-1mm}
\label{tab:cmpr_models}
\end{table*}

\subsection{Measuring circular trajectories}
\label{sec:experimental_setup_dynamics}

We define a circular word trajectory as one where the rotation angle increases linearly with the elapsed proportion of a word segment (constant angular velocity) and completes nearly one full rotation by word offset. (Since frames are discrete, we expect a small gap between word offset and word onset, even for perfect circular dynamics.) 
We benchmark observed dynamics against this optimal reference using two complementary criteria: $R^2$ measures 
the linearity of the relationship between the elapsed proportion of a word segment and the rotation angle, 
and $\mathrm{completion}\%$ measures how complete the rotation is by word offset.\footnote{Strictly speaking, our measures do not distinguish between circles and  some spirals (those with constant angular velocity and one rotation per word). We leave to future work the question of whether this distinction matters.}

Concretely, given a centroid feature $\bar{x}_t \in \mathbb{R}^N$, we finds its two-dimensional projection on the temporal subspace $h_t \in \mathbb{R}^2$.
We define the rotation angle at timestep $t$ relative to the word onset as $\theta_t = \text{arctan2}(h_t) - \text{arctan2}(h_1)$.
We then unwrap the sequence $\theta_1, \ldots, \theta_T$ to track accumulated rotation distance beyond $2\pi$.
Let $d_t \in [0,1]$ denote the elapsed word proportion at timestep $t$.
We fit a linear regression model without an intercept to predict $d_t$ from $\theta_t$. The resulting $R^2$ measures the strength of the linear relationship between rotation angles and elapsed proportion of word segments. The slope of the regression model is then used to estimate the expected rotation angle at word offset as $\hat{\theta} = 1/|\text{slope}|$. Finally, we define $\mathrm{completion}\% = \hat{\theta}/2\pi \times 100$, which measures the percentage of a full rotation completed by the word offset.

\subsection{Experimental details}

We evaluated the word temporal subspaces of the ten speech models in Table~\ref{tab:cmpr_models}. These include
six non-causal Transformer-based self-supervised models using the same architecture but a range of objective functions; two other self-supervised models with different causal architectures and positional encodings; and two ASR models. 
As baselines, we evaluated the untrained HuBERT encoder (used by the first six models), the untrained Whisper encoder, and log Mel features.

For each model layer, we estimated the word temporal subspaces for the Librispeech dev-clean dataset \cite{Panayotov2015-xr} using word boundaries from Kaldi forced alignment \cite{Povey2011-pj}, filtering to include only word segments with durations of 100-700 ms.

\subsection{Results and discussion}
\label{sec:rotation_hubert}

Fig.~\ref{fig:dynamics:word_2d} illustrates our reproduction of Zhang et al.'s results \cite{Zhang2026-ts}, showing that in the temporal subspace estimated by CPCA, average
word trajectories in HuBERT follow a roughly circular shape, 
with the radius of the circle corresponding to word duration. For all word durations, the representations complete nearly one full cycle from word onset to offset.
Across all duration groups, there is a 
strong linear correlation ($R^2 =  0.983$) between the angle of rotation and the elapsed proportion of the word.\footnote{Here, angles are computed with respect to the centroid of each duration group separately. This is impractical for word segmentation, but using one centroid for all groups works for segmentation, as we show in Section~\ref{sec:segmentation}.}

\begin{figure}[]    
    \centering
    \begin{subfigure}[c]{0.85\linewidth}
        \includegraphics[width=1\linewidth]{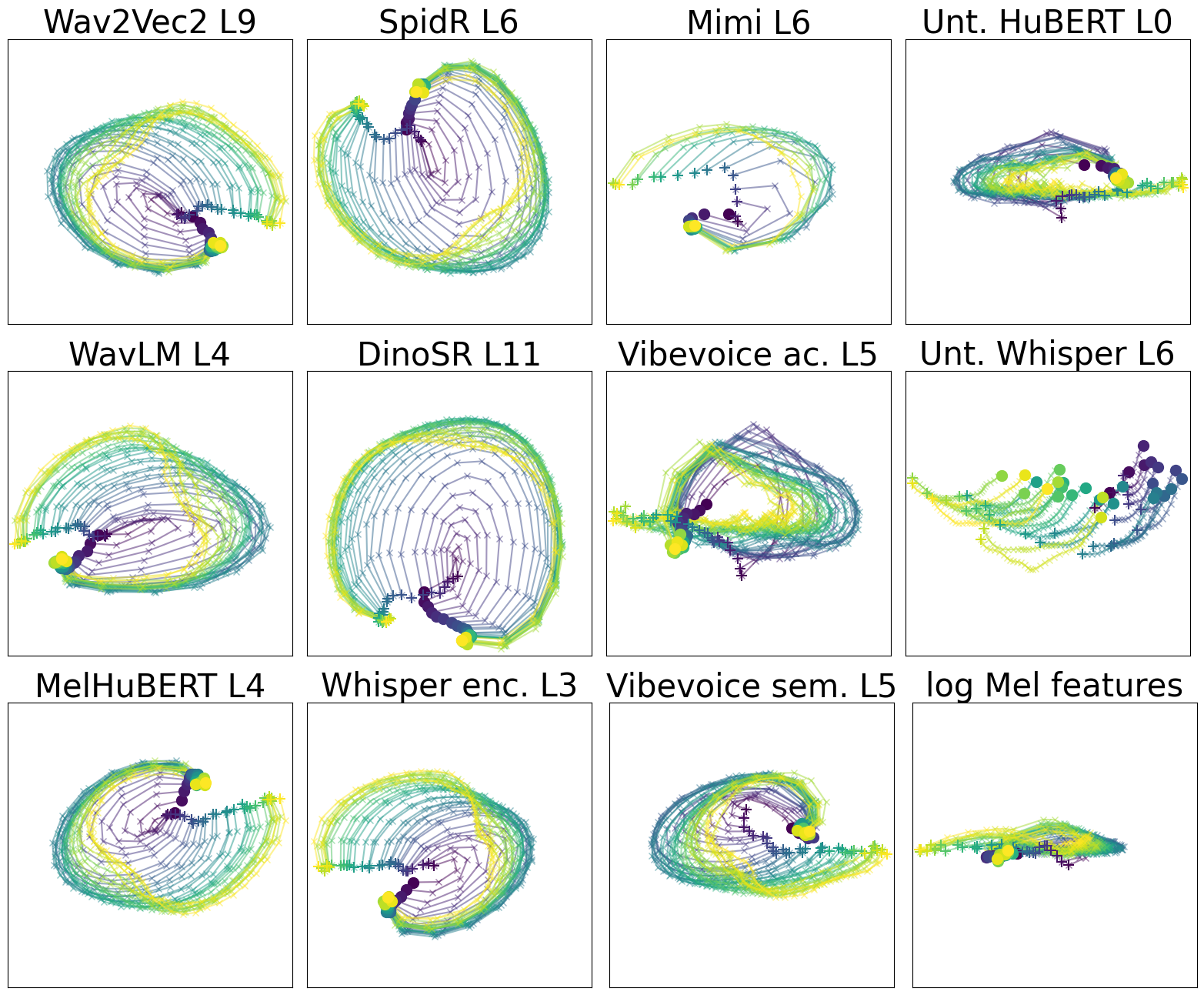}
    \end{subfigure}
    \begin{subfigure}[c]{0.12\linewidth}
        \includegraphics[width=1\linewidth]{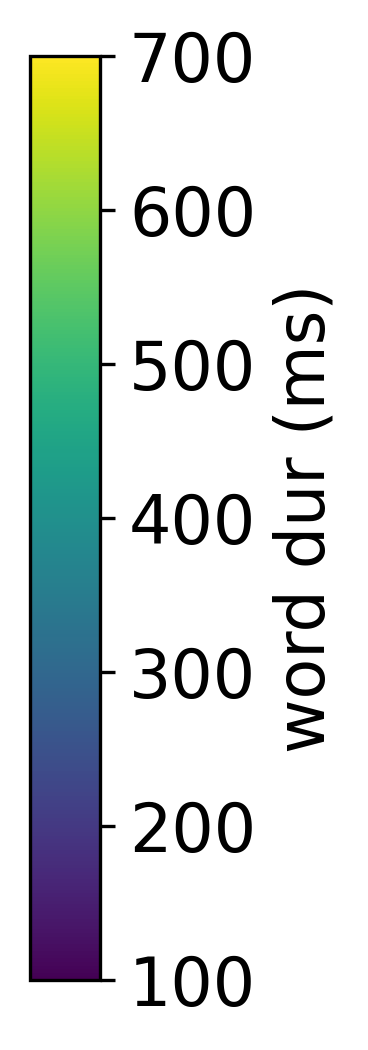}
    \end{subfigure}

    \caption{
    Word trajectories in the top two dimensions of the  CPCA temporal subspace from different model representations, visualized as in Fig~\ref{fig:dynamics}. Rotational dynamics appear in all trained models (first three columns) but only one of the untrained models (last column). For each model, we visualize layer (L) with the best $R^2$ scores. Models in the first two columns are non-causal; those in the third column are causal.}
    \label{fig:more_dynamics}
\end{figure}

\begin{figure}[]
    \centering
    \begin{subfigure}[t]{1.0\linewidth}
         \includegraphics[width=1\linewidth]{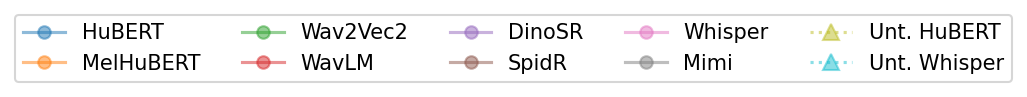}
    \end{subfigure}
    \begin{subfigure}[t]{0.49\linewidth}
         \includegraphics[width=1\linewidth]{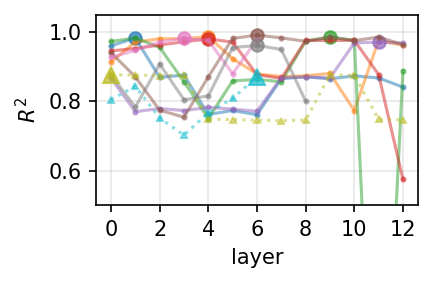} 
    \end{subfigure}
    \begin{subfigure}[t]{0.49\linewidth}
         \includegraphics[width=1\linewidth]{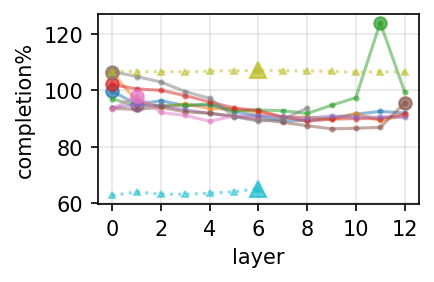} 
    \end{subfigure}
    \caption{
        $R^2$ and $\mathrm{completion}\%$ scores for all layers, where layer 0 indicates the inputs to the Transformer layers. The big markers indicate the layer with the best scores.
        We exclude Vibevoice-acoustic and -semantic as we study only layer 5.
    }
    \label{fig:layer_wise}
\end{figure}

Figs.~\ref{fig:more_dynamics} and~\ref{fig:layer_wise} show that similar rotational dynamics emerge across the other nine trained models we tested, with Fig.~\ref{fig:more_dynamics} visualizing the dynamics of the layer with highest $R^2$, and Fig.~\ref{fig:layer_wise} (left) showing that in all trained models, at least one layer, and often most layers, have an $R^2$ value near 1. Furthermore, all trained models complete roughly one full cycle from word onset to offset (Fig.~\ref{fig:layer_wise}, right), though completion is slightly lower in later layers, indicating a larger gap between average word offset and onset.
Taken together, these results show that circular word trajectories emerge in all trained models and are not due to particular positional encodings or many other architectural 
or training choices.

These findings raise a further question: do these trajectories actually indicate that models represent meaningful knowledge of words acquired through learning? Perhaps such patterns arise in any high dimensional projection of the speech input, without relying on learning. Or, perhaps they are a consequence of models tracking the oscillation of some simple but salient feature of the input (such as intensity) that happens to correlate with words on average. For example, since most English word tokens are monosyllables, tracking the intensity oscillations associated with vowel-consonant alternations might look on average like tracking words.

To begin to answer these questions, we tested representations from the untrained HuBERT and Whisper encoders, as well as log Mel features.
Fig.~\ref{fig:more_dynamics} (final column) reveals that the latter two do not exhibit 
circular trajectories,
demonstrating that not all speech representations (even high-dimensional ones) display this effect. 
On the other hand, the untrained HuBERT representations do show a somewhat circular structure, although with  a flatter shape than the trained models. Fig.~\ref{fig:layer_wise} confirms that the $R^2$ of untrained HuBERT is high, though not as high as the trained models. The cycle completion is also close to optimal, though this model tends to {\em over-complete}, while trained models {\em under-complete} (which is more ``correct'' given the time gap between word offset and onset frames).

Thus, although the average word trajectories reveal quantitative differences between the untrained HuBERT and trained models, there is no clear qualitative difference. 
This motivates us to investigate in more detail the extent to which the observed rotational dynamics arise from model representations that track the time course of individual words, or whether they might arise through tracking syllable-level oscillations, which perhaps could appear to track word cycles under some types of random projection and pooling (e.g., untrained HuBERT). 

\section{Segmenting continuous speech using rotational dynamics}
\label{sec:segmentation}

This section investigates the questions raised above: whether rotational dynamics observed at the averaged-word level are also visible at the instance level, and if so, whether they truly track words, or might instead be tracking syllables.
We focus on the pretrained and untrained HuBERT models, to compare between a model with strong word circles that could arise from learning, and one with weaker circles that could only arise from averaging over unlearned features of the input.

To examine instance-level trajectories, 
we propose Rotary Segmentation (RoSeg), a simple algorithm that segments words using rotational dynamics in temporal subspaces, by marking the end of each rotation as a word boundary. 
We explore two temporal subspace estimators for RoSeg: CPCA (as above) and LEAst-squares Concept Erasure (LEACE; \cite{Belrose2023-ny}), which we find works better for word segmentation.

We find that RoSeg-LEACE achieves strong word segmentation performance using representations from pre-trained (but not untrained) HuBERT, and shows only moderate over-segmentation of multi-syllable words. This indicates that HuBERT contains considerable implicit knowledge of word units, which emerges through learning and much of which is contained in just two dimensions. These results also show that patterns found in average representations may not always translate into meaningful instance-level structure.

\subsection{LEACE for estimating temporal subspaces}

We explore LEACE \cite{Belrose2023-ny} as an alternative to CPCA because previous work suggests that it may be a better concept estimator \cite{Naowarat2026-lt}. 
LEACE was originally proposed as a concept-erasure method that removes linearly decodable information about a target concept from representations while minimally changing the representations \cite{Belrose2023-ny}. Naowarat et al. \cite{Naowarat2026-lt} show that the subspace targeted for erasure can also be interpreted as the subspace associated with the target concept, making LEACE useful for concept subspace estimation. They further show  that, relative to CPCA, LEACE can estimate concept subspaces that capture more information about the target concept.

\begin{figure}[]
    \centering
        \includegraphics[width=8cm]{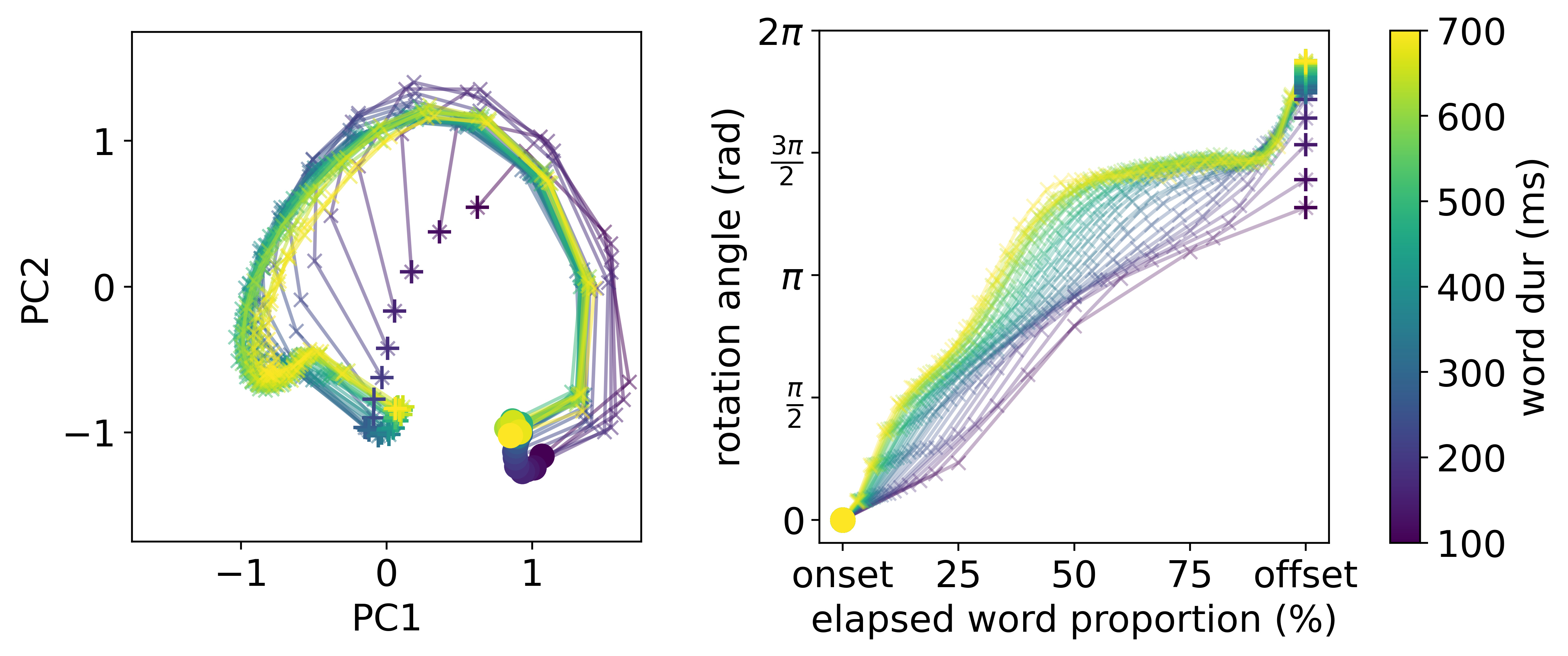}
    \caption{
    Average word trajectories in the top two dimensions of the temporal subspace estimated by LEACE from HuBERT L9 features. The visualization is on the whitened feature space.
    Unlike in CPCA (Fig~\ref{fig:dynamics:word_2d}), words are not binned by duration when estimating the subspace, only for visualization.
    } 
    \label{fig:dynamics_leace}
\end{figure}

In contrast to CPCA, LEACE whitens the features before estimating the subspace. This preprocessing reduces the influence of raw variance scales and feature correlations, allowing variation between average frame centroids to better reflect word-timing information. LEACE estimates the temporal subspace from the cross-covariance between whitened features and word-timing labels (without using word duration information). 
Results on HuBERT are illustrated in Fig.~\ref{fig:dynamics_leace}.

\begin{figure}
    \centering
    \begin{subfigure}[]{1.0\linewidth}
     \includegraphics[width=1\linewidth]{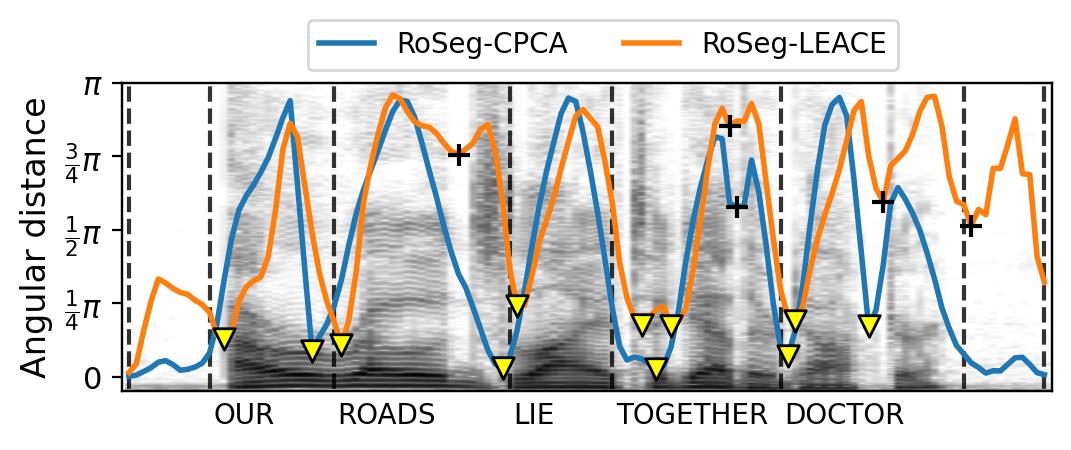}
    \end{subfigure}
    \caption{
    Examples for RoSeg with temporal subspaces estimated by CPCA (blue) and LEACE (orange), overlaid on spectrogram.
    The oscillating lines depict angular distance 
    between the features at each timestep and the first frame of the utterance. 
    RoSeg predicts word boundaries at troughs that fall below an absolute threshold (triangles), and filters out other troughs (pluses).
    True word boundaries are shown as dashed lines.
    }
    \label{fig:oscillation}
\end{figure}

\subsection{Rotary Segmentation (RoSeg)}

RoSeg segments words by assuming that within the temporal subspace, rotational dynamics are continuous and smooth throughout an utterance. That is, we expect to see one cycle per word segment, 
producing a multi-loop trajectory whose cycle completions can be used to identify word boundaries, as illustrated in Fig.~\ref{fig:dynamics:instance}.

Concretely, given 
the frames $u_1, \dots, u_T$ of an utterance,
we first project $u_t$ onto a temporal subspace to obtain temporal features, 
$h_1, \dots, h_T$. 
We take $h_1$ as the reference point and compute the angular distance $a_t$ to each subsequent frame with respect to the origin (i.e., the centroid of the temporal variance), 
$a_t = \text{arccos}( \frac{h_t \cdot h_1}{\lVert h_t \rVert \lVert h_1 \rVert} ) \in [0, \pi]$.
\footnote{
In contrast to the unwrapped rotation angle ($\theta_t$) in Section~\ref{sec:emergent_dynamics}, $a_t$ measures the shortest distance to the reference point in either direction, i.e., $a_t \leq\pi$.
}
The resulting sequence $a_1, \dots, a_T$ 
exhibits an oscillating trajectory over the utterance, as illustrated by the blue and orange lines in Fig.~\ref{fig:oscillation}.

To find word boundaries, we use the prominence-based peak detection algorithm implemented in scipy.signal.find\_peaks, inverting the input to find troughs rather than peaks. 
The same algorithm has also been used for unsupervised word segmentation \cite{Pasad2023-zm, Malan2024-eo}, with the difference being the source of the oscillating signal. In \cite{Pasad2023-zm, Malan2024-eo}, that signal was computed as the cosine distance between each pair of adjacent (mean- and variance-normalized) frame representations. We refer to this method as Dissim.\ and will use it as a baseline below.

Like \cite{Pasad2023-zm, Malan2024-eo}, we apply time-averaged smoothing before detecting troughs. Additionally, we introduce an absolute threshold to filter out high troughs, which indicate a change in the rotation direction without completing a full rotation.
Fig.~\ref{fig:oscillation} hints that these occur mostly in multi-syllable words, which we investigate below.
After filtering, remaining troughs are marked as word boundaries (triangles in Fig.~\ref{fig:oscillation}).\footnote{We release the code at: \url{https://github.com/burin-n/word_timing_trajectory}.}

\subsection{Experimental setup}

\subsubsection{Data and hyperparameters} 
To reduce temporal subspace overfitting, we follow \cite{Naowarat2026-lt} and use separate data splits for subspace estimation and peak detection hyperparameter tuning. The temporal subspace is estimated from 10 hours of speech sampled from the LibriSpeech train-clean-100 subset, filtered to words of 100-700ms duration. (We did not tune these values.) We then follow \cite{Malan2024-eo} and use HuBERT layer 9 features. We tune the absolute and relative prominence thresholds and smoothing window sizes for peak detection to achieve the best R-value \cite{Rasanen2009-xw} on the dev-clean set.

\subsubsection{Evaluation}

For evaluation, we use a tolerance of 20 ms and report F1 scores on boundaries and on tokens (which requires both boundaries to be detected); 
over-segmentation rate (OS, computed as $N_\text{pred}/N_\text{ref} - 1$ using the number of predicted and reference boundaries); and R-value, which combines recall and OS, and reports the distance from the ideal score of 100\% recall and zero OS.\footnote{Evaluation scripts and alignments: \url{https://github.com/s-malan/evaluation}.}

In addition to reporting overall scores, we also compute average scores for each word length (in syllables). To do so, we group word boundaries by the number of syllables in each word. Boundary predictions are then evaluated only within the time spans of words belonging to each group.\footnote{These scores should not be directly compared to the overall scores, because this method double-counts boundaries shared by adjacent groups. For example, given word-boundary indices $[1,5,7,10]$, where the segment $[1,5]$ is a two-syllable word and the segments $[5,7]$ and $[7,10]$ are one-syllable words, the two-syllable group uses $[1,5]$ as its reference span, while the one-syllable group uses $[5,7,10]$, with the boundary $5$ included in both groups.}

\subsubsection{Baselines}
We compare RoSeg to Dissim., which is a state-of-the-art unsupervised method. Although RoSeg uses supervision to find the temporal subspace, the comparison illustrates how analyzing the structure of the representation space can yield a highly informative two-dimensional subspace that outperforms Dissim., even though the latter uses the full feature space.
Additionally, we compare segmentation performance against gold-standard syllable boundaries to test whether RoSeg tracks word or syllable boundaries.

\subsection{Results and discussion}
\label{sec:segmentation_results}

Table~\ref{tab:libri_segmentation} shows that for pretrained HuBERT, both RoSeg-CPCA and RoSeg-LEACE outperform Dissim.\ on three out of four metrics, with RoSeg-LEACE doing particularly well. This indicates that HuBERT representations contain more word-boundary information than previously shown \cite{Pasad2023-zm},
and highlights rotation through a two-dimensional subspace as a compact geometric structure for word-timing information. 

\begin{table}[]
\centering
\caption{Word segmentation performance of RoSeg on Librispeech dev and test sets (in \%), using pre-trained (SSL) and untrained (Unt.) HuBERT representations. We compare CPCA vs.\ LEACE for subspace estimation, and show ablations without absolute thresholding on the dev set. Results for the unsupervised baseline (Dissim.) were obtained using the code at \url{https://github.com/s-malan/prom-wordseg}.
}
\begin{tabular}{ll@{\hskip4pt}r@{\hskip8pt}r@{\hskip8pt}r@{\hskip8pt}r}
\hline
                        \multicolumn{6}{c}{\bf dev-clean}     \\
Model           & Method                    & F1 ($\uparrow$) & OS ($\rightarrow\!\!0$) & Tok F1 ($\uparrow$)    & R-val ($\uparrow$) \\
\hline
- & Gold syllables & 85.5	& 33.9	& 65.9	& 71.1 \\
\hline
SSL HB           & Dissim. \cite{Pasad2023-zm}                          & 43.9 & \textbf{-3.3}  & 17.0     & 52.6  \\
SSL HB           & RoSeg-CPCA                                       & 49.1 & -12.9 & 20.5     & 57.6    \\
                 & \ $-$ abs threshold                    & 48.4 & -4.8  & 19.9     & 56.6    \\
SSL HB           & RoSeg-LEACE                                      & \textbf{66.0} & -4.6  & \textbf{42.3}    & \textbf{71.2}    \\
                 & \ $-$ abs threshold                    & 58.6 & 28.9  & 31.0     & 56.2    \\
\hline
\hline
Unt.\ HB & RoSeg-CPCA                                       & 12.8 & -54.0 & 1.1      & 34.3    \\
Unt.\ HB & RoSeg-LEACE                                      & 16.5 & -32.5 & 1.8      & 35.0    \\
\hline \hline
\vspace{-5pt} \\
                        \multicolumn{6}{c}{ \vspace{1pt}\bf test-clean}   \\
                         \hline 
SSL HB           & Dissim.                           & 43.9 & \textbf{-2.0}  & 16.9     & 52.4    \\
SSL HB           & RoSeg-CPCA                                       & 48.0 & -11.7 & 19.7     & 56.8    \\
SSL HB           & RoSeg-LEACE                                      & \textbf{64.3} & -4.9  & \textbf{39.8}    & \textbf{69.8}    \\
\hline
\hline
\end{tabular}
\label{tab:libri_segmentation}
\end{table}

Table~\ref{tab:libri_segmentation} also shows that RoSeg performs poorly on untrained HuBERT, even though averaged word segments had roughly circular trajectories. This illustrates that the averaged-word dynamics do not reliably translate to individual words, and underscores the importance of evaluating purported geometric structures at the instance level.

\begin{figure}

    \centering
    \begin{subfigure}[t]{0.9\linewidth}
         \includegraphics[width=1\linewidth]{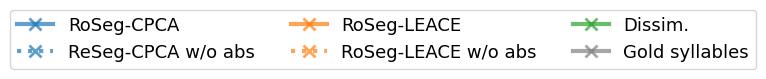}
    \end{subfigure}
    \includegraphics[width=1.0\linewidth]{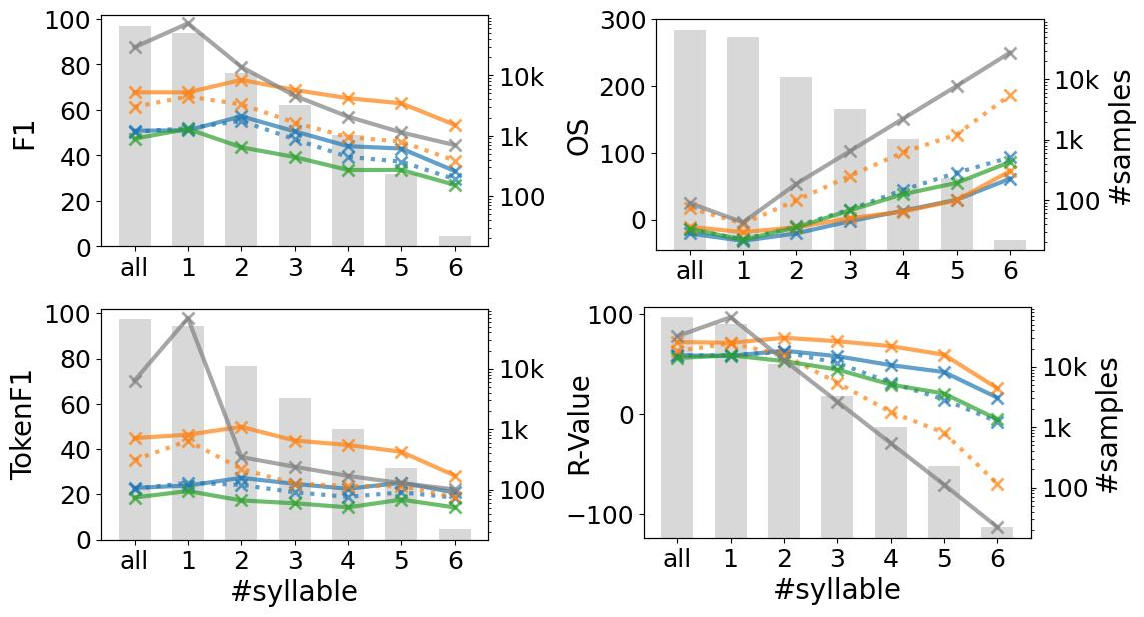}
    \caption{
    Word segmentation performance for words with different numbers of syllables (\#syllable) on the Librispeech dev-clean set, using SSL HuBERT L9 features. RoSeg degrades slower than gold syllables as \#syllable increases, especially on OS and R-value, which account for the number of predicted boundaries. Bars show sample counts (right-hand y-axis).
    }
    \label{fig:multisyllabic_word}
\end{figure}

Finally, we asked to what extent the detected cycles reflect words vs.\ syllables. Table~\ref{tab:libri_segmentation} gives an initial indication, showing that although the gold syllable boundaries yield higher F1 scores than RoSeg (presumably due to the large number of monosyllabic words in English), they also have a much higher OS, indicating that RoSeg is likely finding longer units. 
We confirm this interpretation in Fig.~\ref{fig:multisyllabic_word}, which shows that 
although all methods perform best on monosyllabic words, on longer words RoSeg-LEACE shows considerably better F1 scores than the syllable baseline, and much lower over-segmentation. We conclude that the cycles detected by RoSeg do not simply reflect syllable-level oscillations, such as intensity.

\begin{figure}
    \begin{subfigure}[c]{0.2\linewidth}
        \includegraphics[width=1\linewidth]{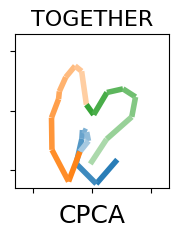}
    \end{subfigure}
    \begin{subfigure}[c]{0.2\linewidth}
        \includegraphics[width=1\linewidth]{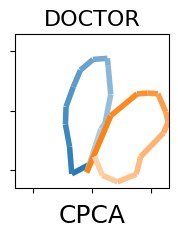}
    \end{subfigure}
    ~~~~
    \begin{subfigure}[c]{0.2\linewidth}
    \includegraphics[width=1\linewidth]{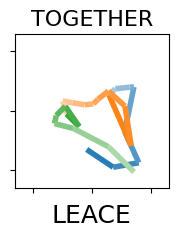}
    \end{subfigure}
    \begin{subfigure}[c]{0.2\linewidth}
    \includegraphics[width=1\linewidth]{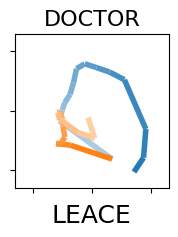}
    \end{subfigure}
    \begin{subfigure}[c]{0.1\linewidth}
    \includegraphics[width=1\linewidth]{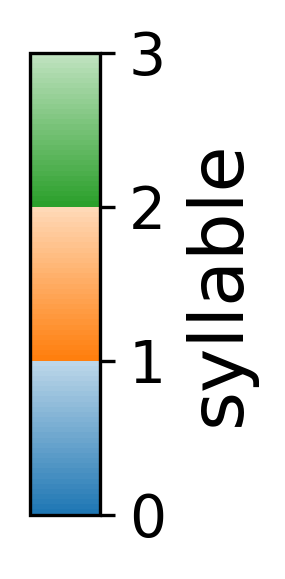}
    \end{subfigure}
    \caption{
    Trajectories of the two multisyllable words from Fig.~\ref{fig:oscillation} through the temporal subspaces estimated from HuBERT L9 by CPCA (left) or LEACE (right). 
    Trajectories are color-coded by syllable identity, with the within-syllable position indicated by a dark-to-light color gradient.
    }
    \label{fig:utterance_syllables}
\end{figure}

On the other hand, our results also suggest that the estimated temporal subspaces may contain some information about syllables as well as words. In particular, results for RoSeg-LEACE without absolute thresholding pattern more like the syllable baseline than the other models do, suggesting that the high troughs could indicate syllable boundaries.

While fully exploring this effect is beyond the scope of this work, examining a few examples revealed intriguing patterns, as shown in Fig.~\ref{fig:utterance_syllables}.
With CPCA, syllables seem to form their own nested loops within the word cycle. These smaller loops may create changes in the speed, but not often the direction, of the overall rotation; thus, the word cycles are less consistent but the absolute threshold has little effect. In contrast, syllables do not form clean loops in the LEACE subspace, but may backtrack along the word cycle. The absolute threshold filters out these cases, leading to better word segmentation.

\section{Conclusion}

In this paper, we introduced a geometry-based approach, motivated by cognitive neuroscience, to analyze word-timing information in deep speech representations. We first showed that circular trajectories tracking average word timing can be found in two-dimensional temporal subspaces across ten deep speech models, suggesting that they may be a general property of neural speech representations. However, a similar but weaker pattern was also found in untrained HuBERT, motivating a further experiment to test whether these average patterns reflect meaningful information at the instance level. We did so by proposing RoSeg, a simple word segmentation algorithm based on rotational dynamics. On pretrained HuBERT representations, RoSeg outperformed an unsupervised baseline, indicating that HuBERT knows more about words than previously demonstrated, and represents the timecourse of individual words as cycles in a low-dimensional subspace. On representations from untrained HuBERT, RoSeg did not succeed,  emphasizing that geometries inferred from average representations may not reflect meaningful instance-level patterns. 
This implies that caution is warranted when interpreting such average geometries (see, e.g., \cite{choi_bd-tp_2026}) and that further investigation would be needed to confirm whether instance-level cycles are present in all models studied here.

More generally, our work contributes to a small but growing set of examples of low-dimensional non-linear manifolds found to represent concepts in deep learning models \cite{csordas_recurrent_2024,Chang2022-ky,AdamYedidia2023-oi} (cf.\ the Linear Representation Hypothesis \cite{park_linear_2023}). Our preliminary evidence that syllable timing may also be encoded in HuBERT representations, together with prior examples of helical structures encoding token position in text models \cite{Chang2022-ky,AdamYedidia2023-oi}, suggests the possibility of a broader theory that rotation may be a general timing encoder, with hierarchical concepts organized as nested rotations. Within speech models, future work could investigate the relationship between word and syllable temporal subspaces, and whether any of these subspaces can be found with weaker supervision.

\section*{Acknowledgment}
We experimented using Claude Code and Codex as coding assistants for code completion, debugging (with and without providing test cases), refactoring (with and without providing detail requirements), and generation (providing only goals). Correctness was reviewed by humans.

\bibliographystyle{ieeetr}
\bibliography{paperpile.bib,references}

\newpage

\end{document}